\documentclass{article}

\usepackage[preprint]{neurips_2022}

\usepackage[utf8]{inputenc} % allow utf-8 input
\usepackage[T1]{fontenc}    % use 8-bit T1 fonts
\usepackage{hyperref}       % hyperlinks
\usepackage{url}            % simple URL typesetting
\usepackage{booktabs}       % professional-quality tables
\usepackage{amsfonts}       % blackboard math symbols
\usepackage{nicefrac}       % compact symbols for 1/2, etc.
\usepackage{microtype}      % microtypography
\usepackage{xcolor}         % colors
\usepackage{natbib}
\usepackage{graphicx}
\usepackage{subfigure}
\usepackage{multirow}
\setcitestyle{square,numbers}
\title{Analyzing Effects of Fake Training Data on the Performance of Deep Learning Systems}

\author{
  Pratinav Seth\thanks{Manipal Institute of Technology, Manipal Academy of Higher Education, Manipal, India}\protect\phantom{\footnotesize 1}\thanks{Authors have contributed equally to this work}\\
  Dept. of Data Science and Computer Applications \\
  \texttt{seth.pratinav@gmail.com} \\
  \And
  Akshat Bhandari\footnotemark[1]\protect\phantom{\footnotesize 1}\footnotemark[2]\\
  Dept. of Computer Science and Engineering \\
  \texttt{akshatbhandari15@gmail.com} \\
  \And
  Kumud Lakara\footnotemark[1]\protect\phantom{\footnotesize 1}\footnotemark[2]\\
  Dept. of Computer Science and Engineering \\
  \texttt{lakara.kumud@gmail.com}\\
}

\begin{document}

\maketitle

\begin{abstract}
  Deep learning models frequently suffer from various problems such as class imbalance and lack of robustness to distribution shift. It is often difficult to find data suitable for training beyond the available benchmarks. This is especially the case for computer vision models. However, with the advent of Generative Adversarial Networks (GANs), it is now possible to generate high-quality synthetic data. This synthetic data can be used to alleviate some of the challenges faced by deep learning models. In this work we present a detailed analysis of the effect of training computer vision models using different proportions of synthetic data along with real (organic) data. We analyze the effect that various quantities of synthetic data, when mixed with original data, can have on a model's robustness to out-of-distribution data and the general quality of predictions. 
\end{abstract}

\section{Introduction}
Deep Neural Networks (DNNs) require large amounts of data to train efficiently and generalize. Computer vision models are always hungry for more data which they use in order to better understand the data distribution and become more accurate. Though data is produced in great volume, it is seldom usable to computer vision models. Large amounts of data in existence may need to be labeled and/or annotated in order to become usable by a computer vision model. All of this is cost intensive especially when the labeling and annotation need to be done by a human in the loop. 

More recently, novel generative deep-learning techniques such as Generative Adversarial Networks (GANs) have demonstrated the ability to create high quality synthetic datasets \cite{karras2017progressive}. 
%Various synthetic datasets have been introduced such as the SUNCG dataset by Princeton \cite{song2016ssc}, MINOS \cite{savva2017minos}, House3d \cite{wu2018building} and Sintel \cite{butler2012mpi}. 
With the advent of high quality synthetic datasets, various data-related challenges in deep learning and particularly computer vision can be combated. Challenges such as class imbalance and robustness to distribution shift can potentially be resolved through the use of synthetically generated datasets. 

In this paper we present a study of the effect of synthetic data on a model's performance. By making use of Conditional Generative Adversarial Network (cGAN) \cite{mirza2014conditional}, we create corresponding synthetic datasets for MNIST \cite{LeCun1998GradientbasedLA}, Fahion-MNIST \cite{Xiao2017FashionMNISTAN} and  CIFAR10 \cite{Krizhevsky2009LearningML}.
%and STL10\cite{Coates2011AnAO}. 
We attempt to evaluate the model performance when trained on different compositions of the core dataset based on the ratio of synthetic and real images. Our results show adding only a few synthetic images to the training dataset not only helps to alleviate the problem of class imbalance but also improves model robustness to distribution shift. 
%------------------------------------------------------------------------
%Real-Syn-Img
\begin{figure}[t]
\centering
\includegraphics[width=0.9\linewidth]{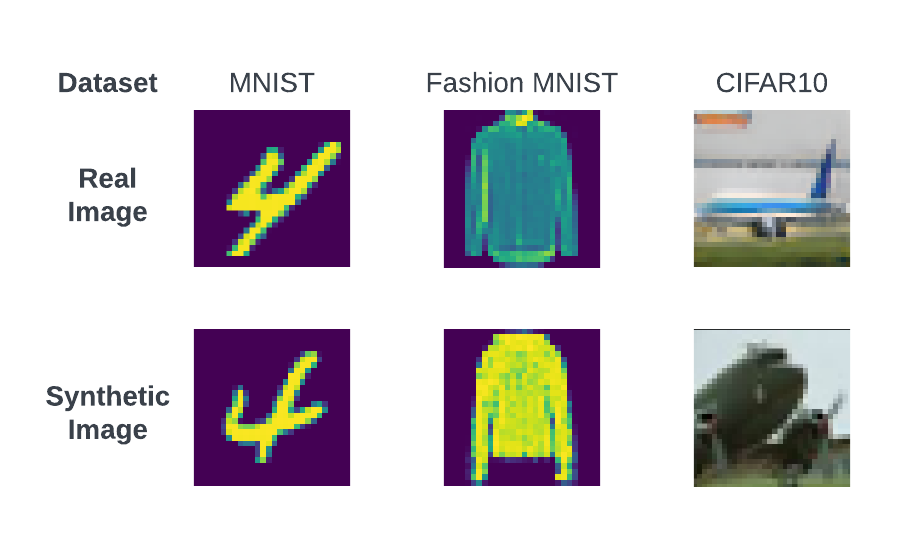}
\caption{Visual Comparison between the Real and Synthetic Images of the datasets used for Experiments.}
\label{fig:twocol}
\end{figure}

\section{Background}
Computer vision tasks are made more difficult by the requirement for large amounts of annotated data. Using low-cost synthetically generated training images is one way to address this problem. This method, however, raises an important question: how should synthetic and real data be combined to optimise model training?
With the growing popularity of GANs, numerous datasets have been introduced in recent years. Some of which include Flying Chairs\cite{Dosovitskiy_2015_ICCV},  FlyingThings3D\cite{Mayer_2016_CVPR}, 
 UnrealStereo\cite{zhang2016unrealstereo}, 
 SceneNet\cite{handa2015scenenet}, SceneNet RGB-D \cite{mccormac2016scenenet}, 
 SYNTHIA \cite{ros2016synthia}, GTAV\cite{richter2016playing}, Sim4CV\cite{muller2018sim4cv}, Virtual KITTI\cite{gaidon2016virtual}, SUNCG dataset by Princeton \cite{song2016ssc}, MINOS \cite{savva2017minos}, House3d \cite{wu2018building} and MPI Sintel \cite{butler2012mpi}.
The use of synthetically generated data for training is one promising approach that addresses the issue of lack of data. However, models trained on synthetic images, frequently suffer from poor generalisation in the real world. Limitations in rendering quality, such as unrealistic texture, appearance, illumination, and scene layout, are common causes of domain gap between the synthetic and real images. 
As a result, networks are susceptible to overfitting to the synthetic domain, resulting in learned representations that differ from those obtained on real images. 

Domain generalisation and adaptation techniques have been proposed to address these issues. \cite{li2017deeper, pan2018two, yue2019domain} Domain adaptation, which aims to tailor the model to a particular target domain by jointly learning from the source synthetic data and the (often unlabeled) data of the target real domain, is frequently used to mitigate the domain mismatch between simulation and the real world. 
Domain generalisation, on the other hand, considers zero-shot generalisation without seeing the target data of real images, and is thus more difficult.

To the best of our knowledge no work so far has tried to combine the synthetic and real images as a single dataset for classification task and studied the effect that their relative ratio can have on model performance and robustness. In this work we investigate the effect of complementing a real dataset with synthetic data for classification tasks. We use conditional GAN (cGAN) to generate synthetic data and present the results for model accuracy when trained on various combinations of synthetic and original data.
\section{Method}
\begin{figure}[hbt!]
    \label{fig:figure2}
    \centering
    \subfigure[]{\includegraphics[width=0.465\textwidth]{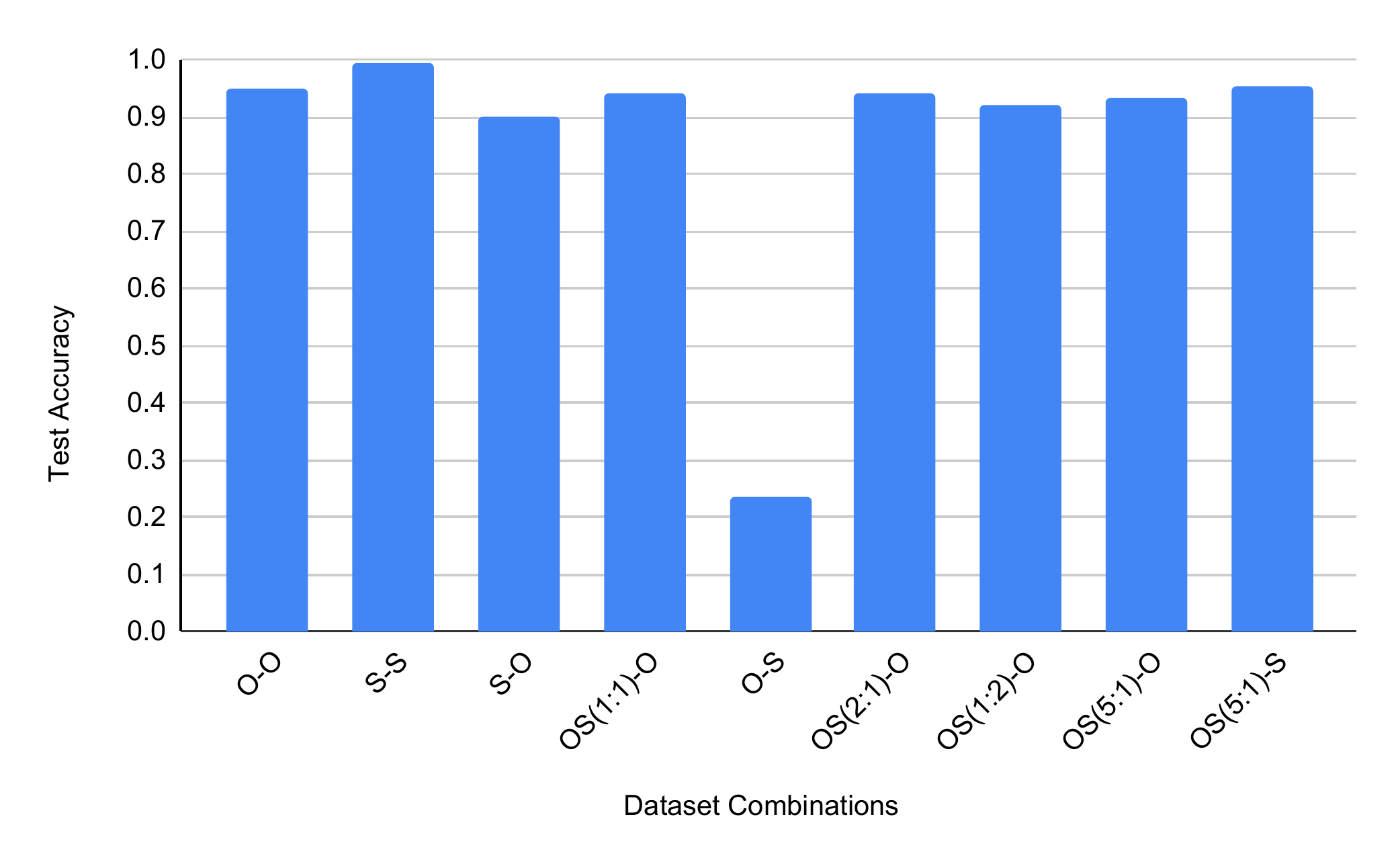}}
    \subfigure[]{\includegraphics[width=0.465\textwidth]{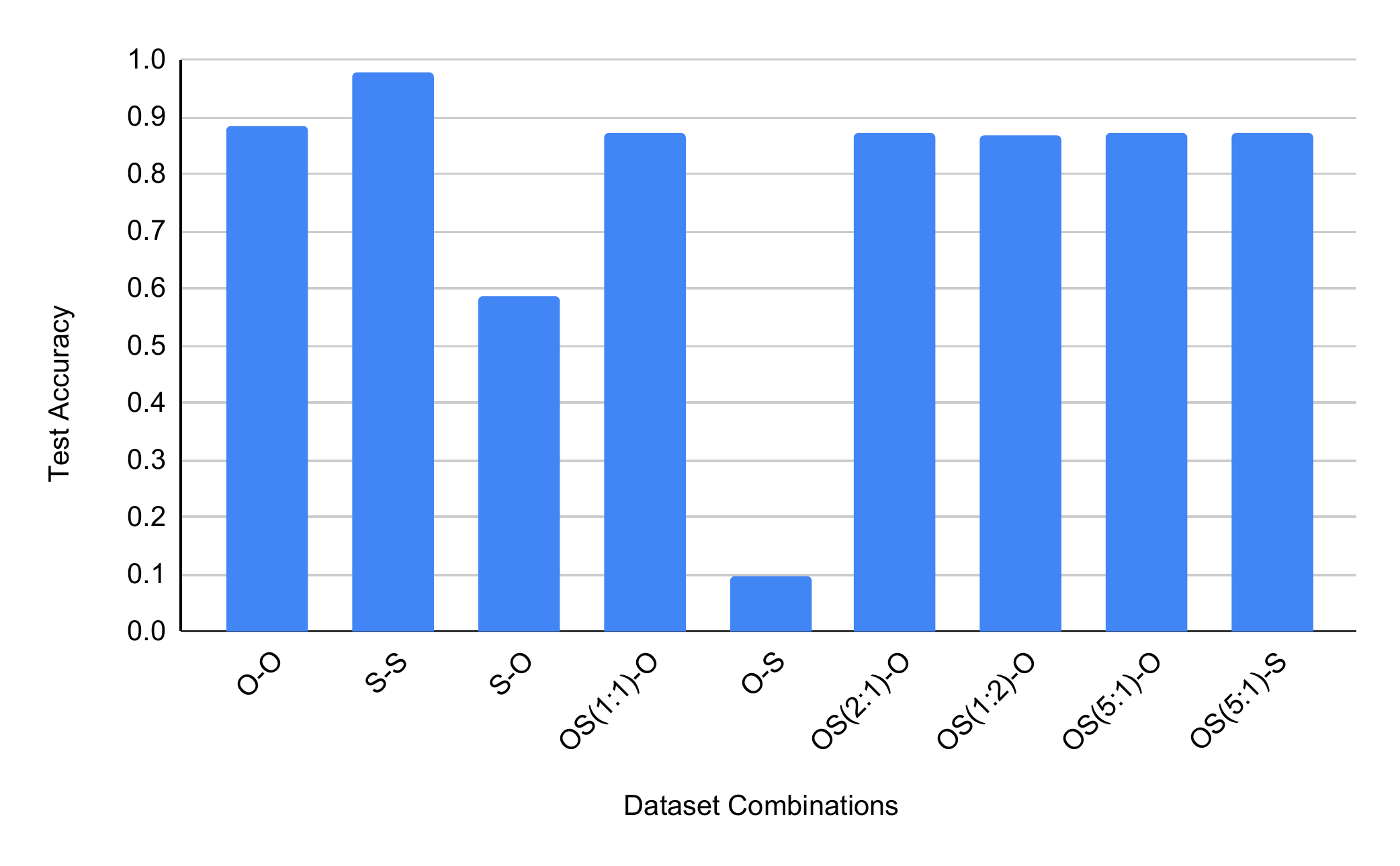}}
    \caption{(a) Test accuracy for different dataset combinations of original and synthetic data for the MNIST Dataset. The X-axis is labeled as training-test set where O: Original, S: Synthetic and OS: Mixed (b) Test accuracy for different dataset combinations of original and synthetic data for the Fashion-MNIST Dataset. The X-axis is labeled as training-test set where O: Original, S: Synthetic and OS: Mixed }  
\end{figure}
We design our experiments in order to evaluate model performance and ability to handle datasets of various domains and complexities to solve the multi-class image classification problem. Models were trained and tested on datasets of constant size throughout the experiments. However, the proportion of original data and synthetically generated data was varied in different ratios. The accuracy of these trained models was calculated on the original organic test split and the synthetically generated test split, respectively.

\subsection{Data Generation}
For our purpose of experimentation, we require real and synthetic data. We focus on the following datasets in this study: 
\paragraph{The MNIST Dataset}\cite{LeCun1998GradientbasedLA} dataset consists of handwritten digits having a training set of 60000 examples and a test set of 10000 examples of grayscale images of size 28x28. The digits have been size-normalized and centered in a fixed-size image. 
\paragraph{The Fashion-MNIST Dataset}\cite{Xiao2017FashionMNISTAN}comprises of 28×28 grayscale images similar to MNIST. It consists of 70,000 fashion products from 10 categories, with 7,000 images per category. The training set has 60,000 images, and the test set has 10,000 images. 
\paragraph{The CIFAR-10 Dataset}\cite{Krizhevsky2009LearningML} is a subset of the Tiny Images dataset and consists of 60000 32x32 color images. The images are labeled with one of 10 mutually exclusive classes. There are 6000 images per class with 5000 training and 1000 testing images per class.
\paragraph{OOD Data}To test the robustness of the models, we use the MNIST-C dataset\cite{Mu2019MNISTCAR}.MNIST-C dataset has corruptions applied to the MNIST test set for benchmarking out-of-distribution robustness in computer vision\cite{Mu2019MNISTCAR}. These corruptions significantly degrade the performance of state-of-the-art computer vision models while preserving the semantic content of the test images\cite{Mu2019MNISTCAR}. We experimented with the Shot noise variant\cite{Mu2019MNISTCAR} for our experiments as it is one of the most common random corruptions that may occur during the imaging process.

We generate synthetic data with the help of conditional GAN (cGAN)\cite{Mirza2014ConditionalGA}. cGANs are a type of Generative Adversarial Networks\cite{Goodfellow2014GenerativeAN} (GANs). %GANs are used for estimating generative models via an adversarial process, in which we simultaneously train two models: a generative model \emph{G} that captures the data distribution, and a discriminative model \emph{D} that estimates the probability that a sample came from the training data rather than \emph{G}\cite{Goodfellow2014GenerativeAN}.
Vanilla GANs \cite{Goodfellow2014GenerativeAN} do not provide control over the class or modes of the data being generated\cite{Mirza2014ConditionalGA}. For our purpose here we require label specific images to be generated. Since we propose using synthetic data in place or in tandem with original data we need to keep the synthetic dataset structure aligned with that of the original data. %hence we use cGAN to generate data as per our data-label based requirements.
cGANs provide a feasible solution by introducing a conditional version of GANs, which can be constructed by simply feeding the label we wish to condition into the generator and discriminator\cite{Mirza2014ConditionalGA}. 

For the purpose of this study we have used StyleGAN2-ada for our experiment in addition to cGAN\cite{Mirza2014ConditionalGA}. We trained cGAN for synthetic data generation to analyze the MNIST and Fashion MNIST datasets. For optimization of both Generator and the Discriminator, we use the Adam Optimizer with a learning rate of $1\times10^{-4}$ for the Discriminator and $2\times10^{-5}$ for the Generator. We use Binary Cross Entropy to measure loss and kept a batch size of 100 images during training. We sample noise from Normal Distribution for noise as input for Generator.
%We use cGAN\cite{Mirza2014ConditionalGA} for generating synthetic data for MNIST\cite{LeCun1998GradientbasedLA} and Fashion-MNIST\cite{Xiao2017FashionMNISTAN} and monitor image quality based on the FID score. 
For CIFAR10 we use StyleGAN2 with adaptive discriminator augmentation(ADA)\cite{Karras2020TrainingGA} to generate synthetic data. We used the official pre-trained class conditional model\cite{Karras2020TrainingGA} trained on CIFAR-10 to produce 32x32 images for our experiments. We monitor image quality based on the FID score.

The different combinations of training and testing data generated and used for the experiments are delineated in table \ref{tab:comb}. 

\subsection{Training Model}
%For the purpose of our experiments we use ResNet18[REF] for the task of image classification. For the cases of MNIST\cite{LeCun1998GradientbasedLA} and FMNIST\cite{Xiao2017FashionMNISTAN}, the model is trained from scratch whereas for CIFAR10\cite{Krizhevsky2009LearningML} we use transfer learning from the imagenet dataset\cite{Deng2009ImageNetAL}.
For our experiments, we use Deep CNN-based model architectures for the task of image classification. For MNIST\cite{LeCun1998GradientbasedLA} and F-MNIST\cite{Xiao2017FashionMNISTAN}, we have used a simple CNN model, while for CIFAR10\cite{Krizhevsky2009LearningML} we used a slightly more complex CNN architecture comprising of 3 convolutional blocks followed by a dense layer and the output layer.  

For each train-test set combination as described in table \ref{tab:comb} we replicate the training procedure in its entirety. Once we perform the experiments with one train-test set combination we conduct the next set of experiments by initializing the model again using the same hyperparameters. 

For MNIST and F-MNIST the models are trained for 10 epochs and for CIFAR10 the models are trained for 50 epochs.
%STL-10\cite{Coates2011AnAO} is a dataset derived from ImageNet having color images of size 96x96. It contains 13,000 labeled images from 10 object classes, among which 5,000 images are partitioned for training while the remaining 8,000 images are for testing. For Generating complex synthetic data, we have used cDCGAN\cite{Zhang2018ImprovingBC}, a conditioned version of DCGAN\cite{Radford2016UnsupervisedRL}. Instead of training GAN with multilayer perceptron as done in cGANs\cite{Mirza2014ConditionalGA}, CNN is used to construct GAN structure in both discriminative and generative models\cite{Radford2016UnsupervisedRL}. This helps them learn complex features and generate images having complex features.
  
%------------------------------------------------------------------------
\section{Analysis}
\begin{table}
\centering
\begin{tabular}{cccc}
    \toprule
    Dataset & Model & Learning Rate & Epoch \\
    \midrule
    MNIST & Simple CNN & $1\times10^{-4}$ & 10 \\
    Fashion-MNIST & Simple CNN & $1\times10^{-4}$ & 10\\
    CIFAR10 & Deep CNN & $1\times10^{-3}$ & 50\\
    \bottomrule
    \\
\end{tabular}
\caption{Different Models and hyperparameters used for different types of datasets for our analysis.}
\label{tab:modelinfo}
\end{table}
Based on the different combinations presented in table \ref{tab:comb} we analyse our results on each dataset.

\begin{table}[]
\centering
\begin{tabular}{llll}
\hline
Dataset                  & Train Data               & Test Data & Test Accuracy \\ \hline
\multirow{9}{*}{F-MNIST} & Original                 & Original  & 0.882         \\
                         & Synthetic                & Synthetic & 0.9771        \\
                         & Synthetic                & Original  & 0.5839        \\
                         & Original                 & Synthetic & 0.0979        \\
                         & Original:Synthetic (1:1) & Orignal   & 0.8699        \\
                         & Original:Synthetic (2:1) & Orignal   & 0.8731        \\
                         & Original:Synthetic (1:2) & Original  & 0.8658        \\
                         & Orignal:Synthetic (5:1)  & Original  & 0.8722        \\
                         & Original:Synthetic (5:1) & Synthetic & 0.8715        \\ \hline
\multirow{9}{*}{MNIST}   & Original                 & Original  & 0.9484        \\
                         & Synthetic                & Synthetic & 0.9926        \\
                         & Synthetic                & Original  & 0.8984        \\
                         & Original                 & Synthetic & 0.2365        \\
                         & Original:Synthetic (1:1) & Orignal   & 0.9409        \\
                         & Original:Synthetic (2:1) & Orignal   & 0.9427        \\
                         & Original:Synthetic (1:2) & Original  & 0.9204        \\
                         & Orignal:Synthetic (5:1)  & Original  & 0.9322        \\
                         & Original:Synthetic (5:1) & Synthetic & 0.9528        \\ \hline
\multirow{9}{*}{CIFAR10} & Original                 & Original  & 0.6771        \\
                         & Synthetic                & Synthetic & 0.7919        \\
                         & Synthetic                & Original  & 0.6523        \\
                         & Original                 & Synthetic & 0.6336        \\
                         & Original:Synthetic (1:1) & Orignal   & 0.6728        \\
                         & Original:Synthetic (2:1) & Orignal   & 0.684         \\
                         & Original:Synthetic (1:2) & Original  & 0.6839        \\
                         & Orignal:Synthetic (5:1)  & Original  & 0.6843        \\
                         & Original:Synthetic (5:1) & Synthetic & 0.7363        \\ \hline
                         \\
\end{tabular}
 \caption{Test Accuracy results for Fashion-MNIST (F-MNIST),MNIST and CIFAR10 using synthetic datasets created by mixing different proportions of real and generated images.}
 \label{tab:comb}
\end{table}
\subsection{MNIST and Fasion-MNIST}

We observe that when the model was trained on just real data, it could not perform well on the synthesized test set. The model failed miserably on a purely synthetic test set. However the model trained solely on synthetic data and tested on the original data still showed relatively better performance. 

We notice that on addition of just a small proportion of synthetic images to the original dataset, the model accuracy started improving. This however can be merely due to the fact that the model was now aware of the synthetic data distribution as well. However the model still shows an impressive revival of accuracy on adding just a few instances of the synthetic data to the training set. Table \ref{tab:fmnist_res} shows the results for the various datasets.

We notice a significant increase in accuracy from the Original-Synthetic(1:1) to the Original-Synthetic(5:1) training-set category as shown in Table \ref{tab:comb}. We further tested the model performance on out of distribution data. For this we test a model trained on different combinations of the datasets on corrupted MNIST. Table \ref{tab:cmnist_res} shows that model accuracy increases when it is trained on a hybrid dataset containing both synthetic and original images. 

Hence models trained on synthetic datasets performed decently on a real dataset. We also observe that when a small amount of synthetic data is included in the training set of the model, it attains similar accuracy levels in both real and synthetic testing sets for both datasets (MNIST and Fashion-MNIST), making it robust to cases where it encounters out-of-distribution images.This is much more prevalent while testing the MNIST model on MNIST-C test set.

%\subsection{CIFAR10}
%\begin{figure}[t]
%  \centering
%    \fbox{\rule{0pt}{2in} \rule{0.99\linewidth}{0pt}}
%   \caption{Comparision of Accuracy of models trained on various proportion of MNIST Original and %Synthetic Dataset}
%   \label{fig:onecol}
%\end{figure}

%We trained a cDCGAN for synthetic data generation to analyze the STL10 datasets. For optimization of both Generator and Discriminator, we have used Adam with a learning rate of  $2\times10^{-4}$. We have used MSELoss, which measures the squared L2 norm between each element in the input and target to measure loss, and kept a batch size of 50 images during training. We sampled noise from Normal Distribution for noise as input for Generator. We have used a pre-trained Resnet18 model using ImageNet weights to analyze the datasets.

%We observed that the model trained on a real dataset could not perform similarly on the synthesized test set. 
%Thus, we can conclude its unable to handle out-of-distribution cases. However, models trained on synthetic datasets can perform better on the real test set. 
%Also, we observed when we included a small amount of synthetic data in the training set. 
%The model attains good accuracy levels in training and performed better in testing sets for both datasets, making it robust to cases where it encounters out-of-distribution images mainly since the dataset contains more complex features than the previous datasets.

\subsection{CIFAR10}
We noticed that the model trained only on  synthetic dataset gives comparable accuracy to the model trained only on original dataset when tested on the original testing split. 
%From this we infer that the synthetic data produced for CIFAR10 lies in a similar domain as the original data. 

In addition to this a subtle yet similar pattern is seen for the cases when the model is trained on just original data and tested on synthetic data and when the model is trained on only synthetic data and tested on original data. The latter shows better performance.

Moreover, for the final combination of O:S (5:1) still shows a favourable trend as with MNIST and F-MNIST. Mixing a small amount of synthetic data improves the model and generally makes them more robust. We use the same number of total training images to train each variant of the model.

%-----------------------------------------------------------------------

\begin{table}[]
\centering
\begin{tabular}{llll}
\hline
Dataset                 & Test Data                 & Train Data               & Test Accuracy \\ \hline
\multirow{11}{*}{MNIST} & \multirow{11}{*}{MNIST-C} & Original                 & 0.9158        \\
                        &                           & Synthetic                & 0.7657        \\
                        &                           & Original:Synthetic (1:1) & 0.8242        \\
                        &                           & Original:Synthetic (2:1) & 0.8572        \\
                        &                           & Original:Synthetic (1:2) & 0.8464        \\
                        &                           & Orignal:Synthetic (5:1)  & 0.9201        \\
                        &                           & Orignal:Synthetic (6:1)  & 0.9566        \\
                        &                           & Orignal:Synthetic (7:1)  & \textbf{0.9596}        \\
                        &                           & Orignal:Synthetic (8:1)  & 0.8923        \\
                        &                           & Orignal:Synthetic (9:1)  & 0.7933        \\
                        &                           & Orignal:Synthetic (10:1) & 0.7625        \\ \hline
                        \\
\end{tabular}
  \caption{Test Accuracy results for MNIST on OOD dataset MNIST-C by models trained using synthetic datasets created by mixing different proportions of real and generated images. }
\end{table}

\section{Results}
We observe a general trend across models and datasets. Table  \ref{tab:comb}
shows the effect that different kinds of synthetic datasets can have on a model's accuracy. We observe that a model trained on a dataset of only original images does not generalize well to out of distribution (and in this case synthetic images) whereas a model trained only on synthetic images does to some extent possess the ability to make correct predictions on original images. 

We also make an important observation that adding only a small number of synthetic images to the original dataset (say in the 1:5 ratio) yields a dataset that is significantly more robust to out of distribution data. The test accuracy of a model trained with such datatsets remains very close to the original/baseline model however the accuracy of such a model on out of distribution images improves by a massive percentage, of the order of 700\% for certain cases. This trend holds for all datasets. 

Synthetic images are generated from random noise and do not possess a definite data distribution hence adding even a small number of such images to the original dataset serves to make it more robust in its prediction. We further verify our experiments on real world out of distribution data using corrupt MNIST. Hence we infer that synthetic data when added to the pure dataset not only serves to alleviate the problem of data shortage but also has secondary effects on the robustness of the model which it helps to improve.

We also observe that though our results hold on simple dataset such as Fashion-MNIST and MNIST, they are not as striking on more complex datasets such as CIFAR10, where classes do not possess many overlapping features and are drastically different from one another.  
 
%------------------------------------------------------------------------
\section{Conclusion and Future Work}
%------------------------------------------------------------------------
In this paper, we performed a detailed analysis of using
an amalgam of synthetic and original data for deep network training. This analysis led to several findings, of which we summarize the most important ones here: (1) A simple model trained only on original data is not as robust to OOD data as compared to a model trained with some synthetic images mixed in (2) while a combination of synthetic and real images benefits models trained for simpler datasets, it might not be so for more sophisticated datasets which possibly contain images with complex overlapping class-wise features.

Since real data is expensive to annotate, the impressive results of synthetic training are valuable. While we focused on datasets of similar complexity, for future analysis, we would like to expand it to different datasets of varying complexity while experimenting with more models and studying the effects of transfer learning. One area of focus will be evaluating various methods with respect to feature overlapping in the dataset.

We hope that our work provides insights on how synthetic images can impact a deep network, pointing the way for future research into developing cost-effective frameworks for training neural networks without needing large amounts of real data.
% \usepackage{multirow}

\bibliography{References}

\newpage

%%%%%%%%%%%%%%%%%%%%%%%%%%%%%%%%%%%%%%%%%%%%%%%%%%%%%%%%%%%%

\end{document}